\documentclass{article}

\usepackage{PRIMEarxiv}

\usepackage[utf8]{inputenc} 
\usepackage[T1]{fontenc}    
\usepackage{hyperref}       
\usepackage{url}            
\usepackage{booktabs}       
\usepackage{amsfonts}       
\usepackage{nicefrac}       
\usepackage{microtype}      
\usepackage{lipsum}
\usepackage{fancyhdr}       
\usepackage{graphicx}       
\graphicspath{{media/}}     

\RequirePackage{amsmath}
\usepackage{changepage}
\usepackage{lscape}
\usepackage{multirow}
\usepackage{multicol}
\usepackage{graphicx}
\usepackage{subcaption}
\usepackage{caption}
\usepackage{lastpage,fancyhdr,graphicx}
\usepackage{epstopdf}
\usepackage{comment}
\title{LPF-Defense: 3D Adversarial Defense based on Frequency Analysis
}

\author{
  Hanieh Naderi \\
  Department of Computer Engineering \\
  Sharif University of Technology \\
  Tehran,Iran\\
   \texttt{h.naderi89@sharif.edu} \\
   \And
  Kimia Noorbakhsh \textsuperscript{\textsection} \\
  Department of Computer Engineering \\
  Sharif University of Technology \\
  Tehran,Iran\\
   \texttt{knoorbakhsh@ce.sharif.edu} \\  
   \And
  Arian Etemadi \textsuperscript{\textsection} \\
  Department of Computer Engineering \\
  Sharif University of Technology \\
  Tehran,Iran\\
   \texttt{arietemadi@ce.sharif.edu} \\
   \And
   Shohreh Kasaei \\
  Department of Computer Engineering \\
  Sharif University of Technology \\
  Tehran,Iran\\
   \texttt{kasaei@sharif.edu} \\
}

\begin{document}
\maketitle

\begin{abstract}
The 3D point clouds are increasingly being used in various application including safety-critical fields. It has recently been demonstrated that deep neural networks can successfully process 3D point clouds. However, these deep networks can be misclassified via 3D adversarial attacks that are intentionality designed to perturb some point cloud's features. These misclassifications may be due to the network's overreliance on features with unnecessary information in training sets. As such, identifying the features used by deep classifiers and removing features with unnecessary information from the training data can improve network's robustness against adversarial attacks. In this paper, the low pass filtered defense framework is proposed to discard the unnecessary information from the training data by suppressing the high-frequency contents in the training phase. The conducted analysis showed that adversarial perturbations place in high-frequency contents of adversarial point clouds. Experiments showed that the proposed defense method achieves the state-of-the-art defense performance against six adversarial attacks on PointNet, PointNet++, and DGCNN models. The findings are practically supported by an expansive evaluation of synthetic (ModelNet40 and ShapeNet) and real (ScanObjectNN) datasets. In particular, improvements are achieved with an average increase of classification accuracy by 3.8\% on the Drop100 attack and 4.26\% on the Drop200 attack compared to the state-of-the-art methods. The method also improves the accuracy of models on the original dataset compared to other available methods. (To facilitate research in this area, an open-source implementation of the method and data is released at \href{https://github.com/kimianoorbakhsh/LPF-Defence}{https://github.com/kimianoorbakhsh/LPF-Defense}.)
\end{abstract}

\keywords{3D deep learning \and frequency domain \and defense \and attack}

\section{Introduction}
\label{sec:introduction}

Point clouds are irregular structures of 3D data, which are used in real-world applications (including healthcare, self-driving cars, drones, robotics, and many more \cite{fernandes2021point, miotto2018deep}). Since point clouds can be receipted directly from scanners, they can precisely capture the shape details. Compared to the 2D counterparts (which are projected forms of 3D data), these irregular structures capture more information from the scene. Therefore, some Deep Neural Networks (DNNs) like PointNet \cite{qi2017pointnet}, PointNet++ \cite{qi2017pointnet1}, and DGCNN \cite{phan2018dgcnn} are designed to be fed by these order-invariant point clouds. One limitation of DNNs is that they are vulnerable to shifting  \cite{xiang2019generating,liu2019extending,hamdi2020advpc,lee2020shapeadv,tsai2020robust,kim2021minimal,liu2021imperceptible}, adding \cite{xiang2019generating,liu2020adversarial,yang2021adversarial,kim2021minimal,arya2021adversarial}, or dropping \cite{yang2021adversarial,wicker2019robustness,zheng2019pointcloud} a small number of points of the original point cloud. These adversarial perturbations lead to model misclassification by distorting some fundamental features of the point cloud. One possible explanation for these shortcomings is the overdependence of deep networks on features with unnecessary information in the training sets, which might not be attending in the object distribution that they attempt to describe. For this reason, identifying the features that deep classifiers use and removing features with unnecessary information from the training data can improve the networks’ robustness against adversarial attacks. 

Some defense methods \cite{xiang2019generating,liu2019adversarial,yang2021adversarial,liu2021imperceptible,kim2021minimal,arya2021adversarial,wen2020geometry,zhou2020lg,dai2021generating,ma2020efficient,tsai2020robust,zheng2019pointcloud,wicker2019robustness} have been proposed to improve robustness in point cloud classification. In general, \emph{modified input} as a preprocess and \emph{modified training} are two common methods to improve adversarial robustness. The adversarial training method \cite{liu2019extending, liang2022pagn} as a \emph{modified training}, trains a model on a mixture of adversarial and clean examples. In fact, they inject redundant feature information into models. The \emph{modified input} method considers preprocessing the input data (before feeding them to the model) to eliminate adversarial perturbations. Removing some redundant feature information is performed by different techniques \cite{xiang2019generating,yang2021adversarial,liu2019extending,zhou2019dup,wu2020if}. 

In terms of adversarial attack and defense in the 2D frequency domain \cite{guo2018low, sharma2019effectiveness, duan2021advdrop, song2021adversarial,wang2020high}, Discrete Fourier Transforms (DFTs) and compression methods are two common methods that can be employed to generate attacks and defends against them.

In terms of adversarial attack and defense in the 3D frequency domain, there are two point cloud attacks that use DFT (graph-based) \cite{liu2022boosting} and \cite{hu2022exploring}, but no defense has been proposed so far. Because of the irregular structure of point clouds, these methods have their own challenges in using imaging methods such as the DFT.

\begin{figure}[!ht]
\includegraphics[width=\textwidth]{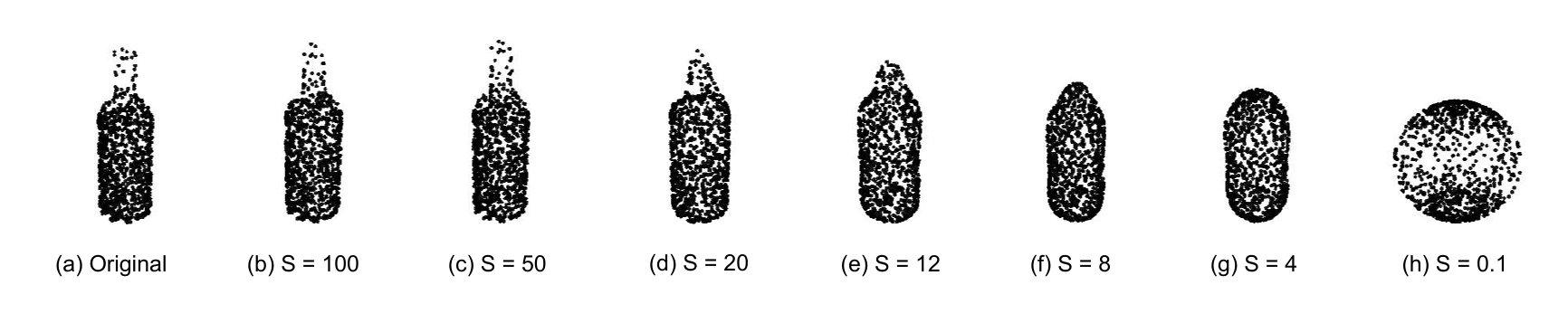}
\caption{ {Illustration of a sample point cloud in different frequencies under spherical harmonic transformation. }
[Lower values of $S$ remove more  3D points with high frequency information, resulting in a more spherical shape.]}
\label{Fig1}
\end{figure}

This paper proposes a defense method from a 3D frequency domain perspective. In the beginning, by using the spherical harmonic functions (to analyze the properties of adversarial point perturbations), it comes to the conclusion that high-frequency components describe fine details of point clouds, whereas low-frequency components show the rough shape of point clouds. An example of such properties can be seen in Fig \ref{Fig1}, which shows a sample point cloud in different frequencies under the spherical harmonic transformation with different high-frequency information. Our analysis shows that most perturbations happen in the high-frequency components for many existing 3D adversarial attacks. Toward this end, this paper proposes a novel adversarial defense method called Low Pass Frequency-Defense (LPF-Defense). The proposed defense first drops the features that contain unnecessary information from the original point cloud and then trains a model on the remaining features, leading to strong robustness against adversarial attacks.

\begin{figure}[!ht]
\includegraphics[width=\textwidth]{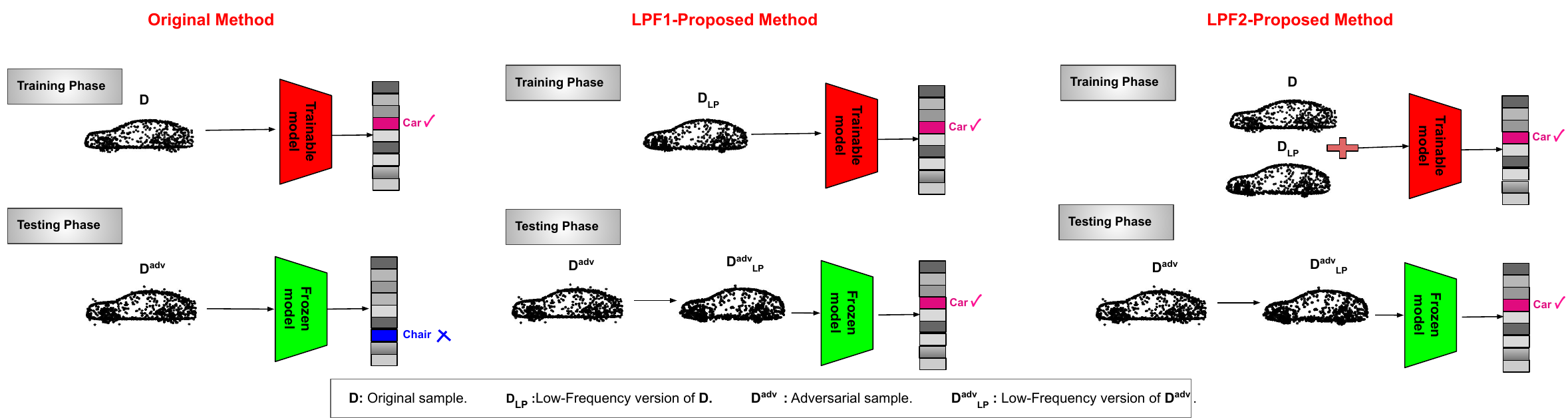}
\caption{ {An overview of proposed (LPF1 and LPF2) defense methods versus original method.}
[Original method fails in robustness against adversarial attacks at test time. Proposed defense methods have overcome this issue.]}
\label{Fig2}
\end{figure}

In other words, the proposed defense method filters out high-frequency input data components as features with unnecessary information when feeding them as the training data. In fact, since the model has never learned the high-frequency components of point clouds, it can be robust against adversarial examples that most of their perturbations occur on high-frequency components. The overall scheme of the LPF-Defense is demonstrated in Fig \ref{Fig2}.

The remaining parts of this paper are organized as follows. Section \ref{sec:Related} introduces the background of the frequency domain, adversarial attack, and defenses on 3D data. The proposed 3D defense method is presented in Section \ref{sec:Proposed}. Experimental results are discussed in Section \ref{sec:Experiment}. Finnaly, Section \ref{sec:Conclusion} concludes the paper.

In summary, the main contributions of this paper are given as follows:
\begin{itemize}
\item Analyzing 3D adversarial examples in terms of frequency domain.
\item Proposing two defense methods based on frequency domain for 3D adversarial examples that outperforme the baselines and state-of-the-art defense methods.
\item Improving the robustness of models on standard inputs by training models using the low-frequency data information.
\end{itemize}


\section{Related Work}
\label{sec:Related}

\subsection{Frequency Methods for Point Clouds}

There are two ways to transform from the spatial domain to the frequency domain; namely, 
the Fourier Transform (FT) \cite{dinesh2020point} and the Spherical Harmonic Transform (SHT) \cite{cohen2018spherical}. In fact, the SHT is the extension of the Fourier series from the circle (one angular component) to the sphere (two angular components). They are a set of orthogonal basis functions defined on the surface of a sphere. Each function defined on the surface of a sphere can be formulated as a weighted sum of these spherical harmonics. It is, therefore, necessary to represent the 3D point cloud on the surface of a sphere. In this regard, Ramsing \emph{et al.} \cite{ramasinghe2020spectral} place the 3D mesh on the unit sphere. Then, rays are cast outward from the shape's center in order to sample the points. They consider the sum of the first and second mesh hit locations as the sample points.

\subsection{Adversarial Attack on Point Cloud}


Adversarial attacks desire to make a classifier to misclassify the input. The attacks and defenses have been analyzed in-depth for 2D data\cite{moosavi2016deepfool,naderi2021generating,carlini2017towards,goodfellow2015explaining,an2013feature,naderi2020scale,madry2019deep}. But, they have been just started to be investigated in the 3D space \cite{xiang2019generating,tsai2020robust,zheng2019pointcloud}. The 3D adversarial attacks are normally classified by different view points as follows:

\begin{itemize}
\item Targeted attacks \cite{xiang2019generating,tsai2020robust} in which the victim model classifies the data point to a specific target class, and untargeted attacks \cite{zheng2019pointcloud} in which the model may classify the point cloud to any class other than the original one. 
\item Point shift (that shifts a few points only and the number of points remains constant) \cite{xiang2019generating} vs. point add (that adds a few points and increases the point numbers) \cite{xiang2019generating,tsai2020robust} vs. point drop (that drops a few points and reduces the point numbers)\cite{zheng2019pointcloud}.
\item On-surface perturbation (perturbations are applied along the object surface.) \cite{xiang2019generating} vs. out-surface perturbation (perturbations are applied outside the object surface; such as noise and outliers) \cite{tsai2020robust,zheng2019pointcloud}.
\item Optimization-based (in which the initial estimate of adversarial perturbation is first considered as an optimization problem, then it is solved using some optimizers.) \cite{xiang2019generating,tsai2020robust} vs. gradient-based (in which gradients of the cost function corresponding to each input point are first acquired and then used to acquire an adversarial perturbation such that the proposed attack has a more tendency towards being misclassified) \cite{zheng2019pointcloud}.
\end{itemize}

There are six different adversarial attacks; namely, Add-CD \cite{xiang2019generating}, Add-HD \cite{xiang2019generating}, Shift-L2 \cite{xiang2019generating}, Shift-KNN \cite{tsai2020robust}, Drop-100, and Drop-200 \cite{zheng2019pointcloud}. They cover all the above categories and are sufficient to examine the ability of defense methods to the extent that they are able to improve the model robustness.




%







ADD-CD, ADD-HD, and Shift-L2 \cite{xiang2019generating} are attacks performed by shifting points with $L_{p}$-norm and adding a limited number of points, clusters, or objects with Chamfer and Hausdorff perturbation criteria.  These criteria force the perturbations to shrink so that they are not noticeable to human eye.

Later, Tsai et al. \cite{tsai2020robust} tried to generate a stronger attack by adding different constraints and proposed a targeted attack, called K Nearest Neighbor (KNN). The KNN attack adds a term to the loss function as a KNN distance constraint (when Chamfer criteria is another constraint that exists to the loss function) so that the points are not too far from the surface and are mostly on-surface perturbations.

Finally, \cite{zheng2019pointcloud} drops the
most critical points based on saliency-based techniques. As such, it calculates a saliency score based on the gradient of the loss function with respect to the input point cloud and drops the points with the highest saliency scores in an iterative manner.

\subsection{Adversarial Defense on Point Cloud}

Given that data quality is one of the most critical issues before performing any analysis, there are several forms of input data modification methods to remove the adversarial noise. The Simple Random Sampling (SRS) \cite{xiang2019generating}, Statistical Outlier Removal (SOR) \cite{zhou2019dup}, saliency map removal \cite{liu2019extending}, and Denoiser and UPsampler NETwork (DUP-Net) \cite{zhou2019dup} are some defense techniques that enhance the model robustness against 3D attacks by applying a preprocessing step before feeding input samples to victim models. These defenses focus on add and shift adversarial examples and cannot perform well on drop attacks. Hence, recent defenses, such as IF-defense \cite{wu2020if}, try to improve the model robustness on all types of attacks. IF-defense \cite{wu2020if} proposed the Implicit Function Defense (IF-Defense) to optimize and restore the input point coordinates by limiting the point perturbation and surface distortion.

In terms of modified training sets, adversarial training \cite{liu2019extending} is one of the most powerful defense techniques. In standard training, the model is trained only on standard point clouds. On the other hand, in adversarial training the model is trained with standard data and adversarial examples. Authors in \cite{liu2019extending} trained models with Shift-l2 attacks and authors in \cite{liang2022pagn} train models with adaptive attacks. In that method, \cite{liang2022pagn} adds different types of attacks to the model. For example, an adaptive attack is designed to cover all types of attacks to improve the model robustness. 

\section{Proposed Method}
\label{sec:Proposed}

\subsection{Problem Formulation}
\label{Problem formulation}

Let $K:\mathbb{R}^{n\times3} \rightarrow \mathbb{R}^l$ be an l-class classifier. Suppose $D\in\mathbb{R}^{n\times3}$ is a spatial representation of a 3D point cloud with $n$ points, $D$ = $\left\{d_{j}| j=1,2, ..., n\right\}$, where each point, $d_{i}$, is represented by its x-y-z coordinates as a vector. Then, $K_{\theta}(D)$ computes the class of input point cloud $D$, which is parameterized by $\theta$. An attacker seeks to achieve the smallest perturbation $\eta$ such that

\begin{equation}
{
\operatorname*{minimize}_\eta \| \eta \| \qquad
\text{subject to}  \quad K_{\theta}(D) \neq K_{\theta}(D+\eta) }
\label{eq:0}
\end{equation}

changes the class of the point cloud with adversarial sample ($D+\eta$), denoted by symbol ($D^{adv}$). The main focus of this paper is to find a solution to improve adversarial robustness of classifiers such that the class of the input point cloud $D$ does not change when corrupted by the perturbation; i.e., $K_{\theta}(D) = K_{\theta}(D^{adv})$. 

The proposed LPF-Defense trains classifiers with the low-frequency version of the input point cloud ($D_{LP}$) (or a mixture of ($D_{LP}$) and ($D$)). The defense filters the high-frequency input data components (as features with unnecessary information) to defeat different types of adversarial attacks. Since the classifiers have learned the low-frequency components of point clouds, they can be robust against adversarial examples that most of their perturbations are on high-frequency components. In order to achieve this motivation, the proposed defense uses spherical harmonic transformation to acquire a low-frequency version when applied to the input point cloud.

\subsection{Low-Frequency Point Cloud Information Extraction}
\label{Low-Frequency Point Cloud Information Extraction}
The purpose of this section is to explain how to extract low-frequency point cloud information, $D_{LP}$, from an input point cloud $D$. 


\textbf{Projecting onto Unit Sphere} The first step is projecting the input point cloud, $D$, onto the unit sphere, centered at its centroid. In fact, each point $d_{i}$ of the point cloud should be converted to the polar coordinate system $(r_{i}, \theta_i, \phi_i)$, where $r$ is the radial coordinate, $\theta\in\mathbb[0, 2\pi]$ is the azimuth angle, and $\phi\in\mathbb[0, \pi]$ is the polar angle.
The projection 

\begin{equation}
D \xrightarrow[\text{}]{\text{project}} f(\theta, \phi) 
\label{eq:11}
\end{equation}

is characterized as a non-negative function $f(\theta, \phi)$ defined on an equiangular sampling grid on the sphere surface, in which $\theta$ and $\phi$ are the co-latitude and longitude, respectively. In more details, each grid point on the sphere surface is equivalent to $r_i$ = $f(\theta_i, \phi_i)$.
The value of $f$ at a particular grid point $(\theta_*, \phi_*)$ is defined as the radius $r_i$ of the point in the point cloud that its polar coordinates $(\theta_i, \phi_i)$ are the closest to $(\theta_*, \phi_*)$. In this projection, several grid points might correspond to the same point in the point cloud. Also, there might be points with no grid points assigned to them.
It is worth noting that the grid needs to be of sufficient resolution to capture the fine details of the input point cloud.
After obtaining the projection on the sphere, the spherical harmonics transform the data to the frequency domain.

\textbf{Spherical Harmonics} Spherical harmonics are a set of complete and orthogonal basis functions for representing any function on the unit sphere $\mathbb{S}^2$. According to~\cite{ramasinghe2020spectral}, any continuous function $f:\mathbb{S}^2 \rightarrow \mathbb{R}$ that satisfies a certain set of conditions can be written as

\begin{equation}
f(\theta, \phi) = \sum_{l=0}^{\infty}\sum_{m=-l}^{l}c_l^mY_l^m(\theta, \phi)
\label{eq:00}
\end{equation}

where $Y_l^m(\theta, \phi)$ are the spherical harmonics basis functions of degree $l$ and order $m$, and $c_l^m$ is the corresponding coefficient of the basis function. The spherical harmonics basis functions are defined as

\begin{equation}
{Y_l^m(\theta, \phi) = N_l^m P_l^m (\cos(\pi))e^{im\theta}}
\label{eq:6}
\end{equation}

in which $N_l^m$ are the normalization coefficient, $P_l^m$ are the associated Legendre function, and $i = \sqrt[2]{-1}$ is the imaginary unit.

The coefficients $c_l^m$ are the frequency domain representations of function $f$ that can be calculated as

\begin{equation}
{c_l^m = \int\limits_0^\pi \int\limits_0^{2\pi} f(\theta, \phi)Y_l^m(\theta, \phi)\sin \phi d\phi d\theta}
\label{eq:3}
\end{equation}

for a given function $f$. In implementations of this paper, the python library "pyshtools"~\cite{wieczorek2018shtools} was used to perform spherical harmonics related operations.

\textbf{Low Pass Filtering in Frequency Domain} After obtaining the point cloud projection onto the sphere, spherical harmonics are used to calculate the frequency domain coefficients.
Each coefficient $c_l^m$ is then multiplied by a corresponding weight $w_l$, to calculate the updated coefficient $\hat{c}_l^m$ as

\begin{equation}
\hat{c}_l^m \xleftarrow[\text{}]{\text{updated coefficient}} c_l^m \times w_l 
\label{eq:33}
\end{equation}

which has the effect of low pass filtering. Weights $w_l$ ($0 \le l \le \infty$) come from a Gaussian distribution $\mathcal{N}(0, \sigma)$, such that $w_{l=0} = 1$.
The higher order coefficients ($c_l^m$'s with higher $l$) are multiplied by a smaller weight, resulting in low pass filtering. This results in diminishing noise and outliers. A low pass version of $f(\theta, \phi)$ is defined by

\begin{equation}
\hat{f}(\theta, \phi) = \sum_{l=0}^{\infty}\sum_{m=-l}^{l}\hat{c}_l^mY_l^m(\theta, \phi)
\label{eq:1}
\end{equation}

which is expanded with updated coefficient $\hat{c}_l^m$.
Our studies showed that the Gaussian filter outperforms the box filter (which are often used to cut-off frequencies). 

\textbf{Reconstructing the Filtered Point Cloud} The final step is to transform the coefficients $\hat{c}_l^m$ back into spatial domain for retrieving the low-passed data $hat{f}:\mathbb{S}^2 \rightarrow \mathbb{R}$.
For each spherical angle pair $(\theta, \phi)$ that have at least one point $d_{i}$ from the point cloud assigned to them during the initial projection, a point is generated based on the new value of $\hat{f}(\theta, \phi)$.
These newly generated points reconstruct the low pass filtered point cloud $D_{LP}$ as

\begin{equation}
D_{LP} \xleftarrow[\text{}]{\text{project}} \hat{f}(\theta, \phi).
\label{eq:4}
\end{equation}

Note that the point cloud size might be reduced after this process. The potential loss in the number of points can be compensated by randomly re-sampling existing points. 
In contrast to~\cite{ramasinghe2020spectral}, point clouds themselves are the primary inputs here and not their corresponding meshes. This is crucial to the semantics of defending against adversarial attacks, as one is only given the point cloud data to perform the defense.

\subsection{Frequency-Based Analysis Attacks}
\label{Proposed:Frequency-Based Analysis attacks}
The issue of defense is raised by introducing adversarial attacks. A better analysis of adversarial attacks gives a better view of how to generate the defense. The adversarial attack retains the original appearance of the object and deceives the model by modifying some points in a way that is not noticeable to the human eye. It is usual to suppose that adversarial perturbations affect more high-frequency components than low-frequency ones. This paper studies this assumption by analyzing the adversarial attacks in the frequency domain. For this purpose, the $\mathrm{Dis}_{\mathrm{c}}$ function measures the average dissimilarity between the spherical harmonics coefficients of the original point cloud and the adversarial one in the data test

\begin{equation}
\mathrm{Dis}_{\mathrm{c}} = \frac{1}{N}\sum_{i=0}^{N}|{\frac{[c_l^m]_{i}^{adv}-[c_l^m]_{i}^{org}}{[c_l^m]_{i}^{org}}|}
\label{eq:5}
\end{equation}

where $N$ is the number of point clouds in the data test, and $[c_l^m]_{i}^{adv}$ and $[c_l^m]_{i}^{org}$ are the coefficients corresponding to the spherical harmonics basis functions in the original point cloud $D$ and the adversarial one $D^{adv}$, respectively. Also, $c_l^m$ stands for the coefficients of Spherical Harmonics Transform. For visualization to perform well under different adversarial perturbations, a normalization process is performed by dividing these differences by the coefficients of original point clouds. According to the results, most adversarial perturbations occur at medium and high frequencies.


\begin{figure}[!ht]
\includegraphics[width=\textwidth]{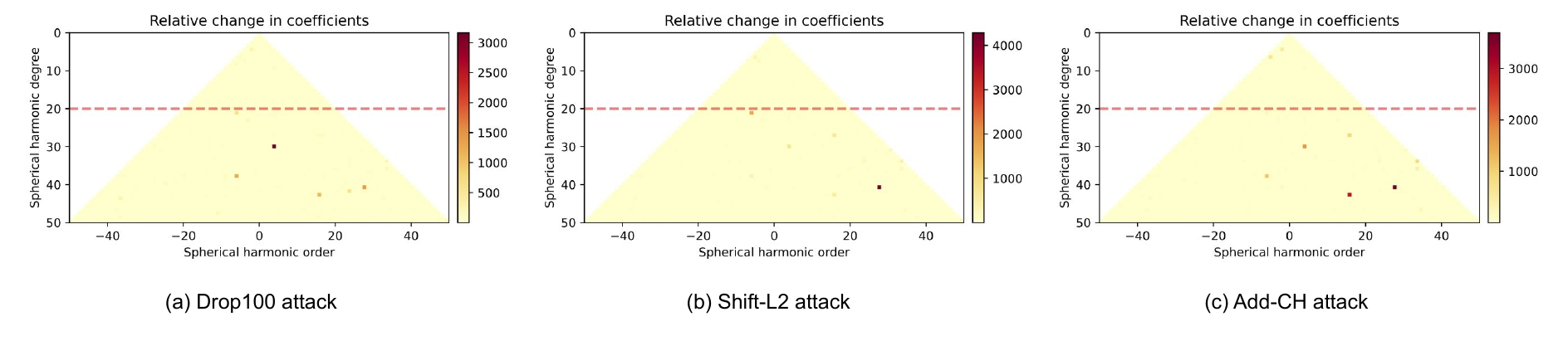}
\caption{{\bf Spherical harmonic coefficients for different adversarial perturbations on ModelNet40. }
Upper vertex and two lower vertices of the triangle represent the lowest and highest frequency components in the frequency space. Results show that more remarkable changes occur in mid- and high-frequency ranges. [Lighter color reveals a more remarkable change for a particular frequency component between original and adversarial point clouds.]}
\label{Fig4}
\end{figure}

\subsection{Adversarial Defense with Low-Frequency Information of Point Cloud}

The two proposed $LPF1$ and $LPF2$ defenses train classifiers with a low-frequency version of the input point cloud $D_{LP}$ and a mixture of $D_{LP}$ and $D$, receptively. In both techniques, classifiers are tested with $D_{LP}$ of adversarial examples.
The general idea behind both proposed defenses is to use a training model that eliminates high-frequency input data components as features with unnecessary information in order to defeat different types of adversarial attacks. The LPF2 method appears similar to the popular adversarial training, but it has some key differences. It focuses on data injection by removing their high-frequency components to achieve a stable model performance. In contrast, adversarial training focuses on data injection by adding some high-frequency features to the model during the training phase, which can cause the instability of the 3D model when facing unseen attacks (other redundant features that the model has not learned during the training phase). Due to this fact, the $LPF2$ method enhances the model's performance over the standard adversarial training.

\section{Experimental Results}
\label{sec:Experiment}

\subsection{Settings}


\textbf{Datasets} The experiments in this paper used aligned benchmark ModelNet40 \cite{wu20153d}, real scanobjectNN\cite{uy2019revisiting}, and ShapeNet \cite{chang2015shapenet} dataset for 3D object classification. The ModelNet40 dataset contains 40 object classes and 9,843 3D Computer-Aided Design (CAD) objects for training, and 2,468 3D CAD objects for testing. The ScanObjectNN dataset is an object dataset from real-world that contains object scans in indoor environments. It has 15 object classes, 11,416 objects for training and 2,882 objects for testing. The ShapeNet dataset used in this paper is a subset of the full ShapeNet dataset. It covers 8  object categories with 12235 training objects and 616 testing objects. (The whole training dataset was not selected due to the high computational cost of generating the attacks. These 8 objects include the following categoris: Display, Chair, Cabinet, Bag, Sofa, Pillow, Shelf, and Bed.)

\textbf{3D Models} Three state-of-the-art models, including the PointNet \cite{qi2017pointnet}, PointNet ++ \cite{qi2017pointnet1}, and DGCNN \cite{phan2018dgcnn} are adopted as victim classifiers that run on the ModelNet40 dataset. The models are trained with default settings. (The experiments were performed on a machine equipped with one NVIDIA Tesla P100-PCIe and 16 GB memory.) 

\textbf{Attack Settings} The proposed method has been tested on un-targeted/targeted attacks. Attacks are fed to the pre-train victim model to evaluate the accuracy. Attacks are re-produced for each model for a fair comparison between the LPF-Proposed and the base defense methods according to IF-Defense \cite{wu2020if} settings.
A target class, which is not equal to the ground-truth class, has been randomly assigned to each data test for targeted adversarial attacks. This assignment of the target classes was maintained unchanged in all attacks to remove the randomness effect. Therefore, there are 2468 attack pairs (victim, target) to measure the accuracy.
For un-targeted attacks, all test objects contain 2,468 objects fed to the model to estimate the accuracy. Therefore, the basic adversarial examples including the point shifting (Shift-L2) \cite{xiang2019generating}, the point adding (Add-CD and Add-HD) \cite{xiang2019generating}, the kNN attack (Shift-kNN) \cite{tsai2020robust}, and the point dropping (Drop-100 and Drop-200) \cite{zheng2019pointcloud} were employed for performance comparison purposes. When Shift-L2, Add-CD, Add-HD, and kNN attacks optimized a Carlini \& Wagner ($C\&W$) function with L2-norm, Chamfer, Hausdorff, and both Chamfer and K-nearest neighbors distance as a perturbation metric, respectively. The same as \cite{wu2020if}, a 10-step binary search with 500 iterations in each step was utilized to generate the Shift-L2, Add-CD, and Add-HD attacks. Also, 2500 iterations were used for Shift-KNN. Furthermore, for add points attacks (ADD-HD and ADD-CD), 512 points were added to 1024 points in each point cloud.
Drop-100 and Drop-200 attacks removed 100 and 200 points from 1024 points with the highest saliency scores, wherein every iteration, 5 points with the highest saliency scores were dropped. Then, a new saliency map was constructed for the remaining points. This process was repeated in next iterations to drop 100 and 200 points.
Point dropping was under un-targeted settings and others were under targeted settings. All experiments in this paper were implemented using PyTorch. 

\textbf{Defense Settings} To verify the validity of the proposed defense method, this method was compared with the SRS \cite{yang2021adversarial}, SOR \cite{zhou2019dup}, DUP-Net \cite{zhou2019dup}, If-Defense \cite{wu2020if}, Adv Training with Shift-L2 \cite{liu2019extending}, and Adv training with PAGN \cite{liang2022pagn} baselines. It is noteworthy that “adversarial” is simplified with “adv” in the text and tables.
In SRS, the number of dropped random points was 500. In SOR, the hyperparameters were set to k = 2 and $\alpha$ = 1.1. 
If-Defense suggested three different versions. This paper reports the results of IF-Defense based on optimization with ConvONEt implicit function networks, which is the best version of the If-Defense.
The rest of the settings were in accordance with related reference papers.

\subsection{Effect of Fourier and Spherical Harmonic Transformations}
\label{Comparison of low-frequency information with Fourier and Spherical harmonic transformation}

\begin{figure}[!ht]
\includegraphics[width=\textwidth]{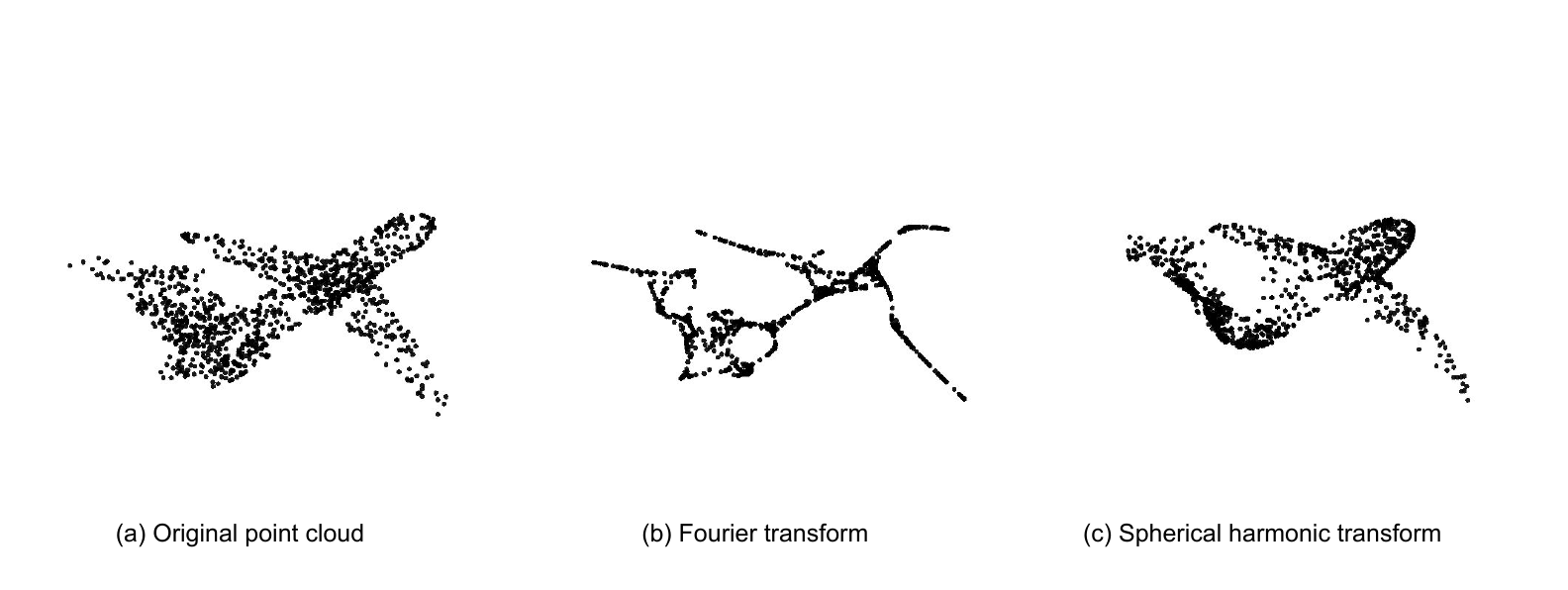}
\caption{ Comparison of low-frequency point cloud information of a random ModelNet40 sample.
[Spherical harmonic transformation removes corners and preserves point clouds' uniform distribution. In contrast, Fourier transformation extracts a skeleton of point clouds very narrowly.]}
\label{Fig3}
\end{figure}

Since the general purpose of this paper is to train the model with low-frequency information, both transformations are analyzed on model training. Fig \ref{Fig3} illustrates extracting the low-frequency content of a sample of ModelNet40 data with FT and SHT. As shown in Fig \ref{Fig3}, teh SHT removes corners and preserves point clouds' uniform distribution. In contrast, the FT extracts a skeleton of point clouds very narrowly. Due to the fact that points in the low-frequency version of data based on the FTare concentrated in certain regions and are not uniformly distributed, they gained less model performance than the SHT. This means that the uniform distribution of points on the surface of the point cloud has a high effect and therefore is essential in the training phase. In addition, the proposed LPF-Defense method seeks to eliminate high frequencies where attack perturbations probably are more concentrated, such as corners. Therefore, for the purpose of this paper, it seems more appropriate to use the SHT.


\subsection{Analysis of Adversarial Attacks in Spherical Harmonic Transformation}
\label{Experiment_Adversarial attack analysis in frequency domain}

This part analyzes the effect of adversarial attacks on frequency components by the $\mathrm{Dis}_{\mathrm{c}}$ function that was introduced in Section \ref{Proposed:Frequency-Based Analysis attacks}. The distribution of perturbations (point shifting (Shift-L2) \cite{xiang2019generating}, the point adding (Add-CD) \cite{xiang2019generating}, and the point dropping (Drop-100) \cite{zheng2019pointcloud}) in the frequency domain is visualized in Fig \ref{Fig4}. The top vertex of the triangle shows the coefficient of low-frequency components, and as it moves down, the coefficient of high-frequency components are displayed. Based on the results, most of the adversarial perturbations are found in the mid- and high-frequency components of attacks. In other words, adversarial attacks mostly deceive the model by modifying high-frequency components. 
According to Fig \ref{Fig4}, the frequency components in the 20 rows above the triangle ($S=20$) change slightly and most of changes happen below that. This has happened  in almost every three adversarial attacks.
An approximate cut-off frequency range, $S=20$, is now available for finding low-frequency versions.

\begin{figure}[!ht]
\includegraphics[width=\textwidth]{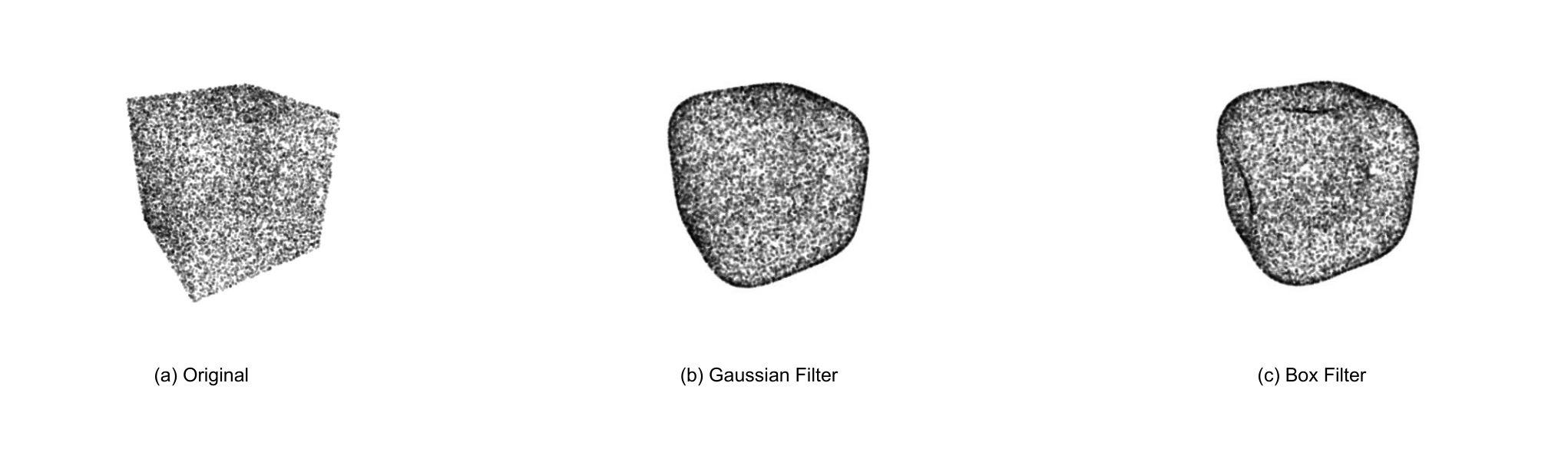}
\caption{{Effect of low pass filtering with Gaussian filter and Box filter. }
(a) Original point cloud. (b) Gaussian low-pass-filtered with $S$=5. (c) Box filter, when all frequency components of degree more than 5 are eliminated. [Gaussian filter performs a more natural blurring effect than box-like filter, when ripples are seen in the flat regions on (c).]}
\label{Fig5}
\end{figure}

\textbf{Cut-off Frequency}
Cut-off frequencies can be performed in two ways of Box filter or Gaussian filter. Box filtering means, from one frequency onwards, all components are discarded (these frequencies would be set to zero). Gaussian filter means frequency components are weighted based on the Gaussian distribution (These frequencies would be high near zero and then decrease at higher frequencies according to the shape of decay of the Gaussian distribution.) By setting the standard deviation, $S$, the Gaussian filter can control the cut-off frequencies. In other words, the higher $S$ gets, the higher the cut-off of high-frequencies occur. In Fig \ref{Fig5}, the results of two types of low pass filtering are observed. The ripple in the flat region is due to the frequency cut-off with the Boxing filter. If a Gaussian filter is used (instead of the Boxing filter), this ripple will be removed.
The impulse response of a Box filter is the sinc function. That is why ripples are seen in the flat regions (flat surfaces of cubes seem to bend in after filtering). Note that Gaussian in the time domain maps to Gaussian in the frequency domain and does not suffer from ripples. Consequently, there will be no bending-in on the flat surfaces.


\begin{table}
\caption{
 {Comparison of classification accuracy of proposed defenses with other defense strategies, under various attacks on PointNet and ModelNet40 datasets.}}

\centering
\tiny
\label{table:PointNet}
\begin{tabular}{llllllll}
\toprule

\bf Defenses  & \multicolumn{7}{|c}{\bf Attacks}\\
\midrule
& Clean & Shift-L2 \cite{xiang2019generating} & Add-CD \cite{xiang2019generating} & Add-HD \cite{xiang2019generating} & Shift-KNN \cite{tsai2020robust} & Drop-100 \cite{zheng2019pointcloud} & Drop-200 \cite{zheng2019pointcloud}\\
 
No-defense & 88.39\% & 0.00\%&0.00\%&0.00\%&8.55\%&60.02\%&33.03\%\\
 
SRS \cite{yang2021adversarial} & 87.35\% & 77.65\% & 76.58\% & 73.27\% & 58.52\% & 63.82\% & 38.92\%\\
 
SOR \cite{zhou2019dup} & 87.83\% & 82.75\% & 82.63\% & 82.46\% & 76.58\% & 64.33\% & 43.05\% \\
 
DUP-Net \cite{zhou2019dup} &  87.69\%& 84.49\%&83.80\%&82.09\%&80.28\%&67.41\%&46.83\%\\
 
If-Defense \cite{wu2020if} & 84.36\%& 86.31\%&86.84\%&86.73\%&\textbf{82.74\%}&77.74\%&66.97\%\\
 
Adv Training (Shift-L2) \cite{liu2019extending} & 88.18\% & 43.28\%&49.35\%&53.47\%&39.22\%&70.23\%&65.79\%\\
 
Adv Training (PAGN) \cite{liang2022pagn} &87.01\%& 84.83\% &61.75\%& 64.35\%& 65.46\%&66.29\%&49.61\%\\
 
LPF1-Proposed & 84.87\% &84.97\% &85.13\% &84.85\%&83.59\%&77.10\%&61.16\%\\
 
LPF2-Proposed &\textbf{91.78\%}& \textbf{86.91\%}&\textbf{87.32\%}&\textbf{86.99\%}&86.02\%&\textbf{81.64\%}&\textbf{68.52\%}\\
 
\bottomrule
\end{tabular}

\end{table}

\begin{table}
\caption{
 {Comparison of classification accuracy of proposed defenses with other defense strategies, under various attacks on PointNet++ and ModelNet40 datasets.}}
\tiny
\centering
\label{table:PointNet++}
\begin{tabular}{llllllll} 
\toprule

\bf Defenses  & \multicolumn{7}{|c}{\bf Attacks}\\
\midrule
& Clean & Shift-L2 \cite{xiang2019generating} & Add-CD \cite{xiang2019generating} & Add-HD \cite{xiang2019generating} & Shift-KNN \cite{tsai2020robust} & Drop-100 \cite{zheng2019pointcloud} & Drop-200 \cite{zheng2019pointcloud}\\
 
No-defense & 89.58\% & 0.00\%&7.29\%&6.32\%&0.00\%&80.28\%&68.89\%\\
 
SRS \cite{yang2021adversarial} & 83.71\% & 74.01\%&65.28\%&43.12\%&49.89\%&64.53\%&40.05\%\\
 
SOR \cite{zhou2019dup} & 87.02\% & 77.64\% &72.93\%&72.4\%&61.42\%&74.09\%&-69.28\%\\
 
DUP-Net \cite{zhou2019dup} & 85.72\% & 81.01\%&75.78\%&72.46\%&74.81\%&76.41\%&72.10\%\\
 
If-Defense \cite{wu2020if} & 88.97\% & \textbf{86.97\%}&80.21\%&76.15\%&85.59\%&84.61\%&78.99\%\\
 
Adv Training (Shift-L2) \cite{liu2019extending} & 89.14\% & 20.45\%&13.01\%&10.12\%&9.05\%&80.51\%&66.98\%\\
 
LPF1-Proposed & 76.17\% & 83.31\%&80.35\%&74.03\%&83.59\%&72.16\%&66.53\%\\
 
LPF2-Proposed & \textbf{90.92\%} & 85.90\%&\textbf{83.43\%}&\textbf{77.23\%}&\textbf{86.55\%}&\textbf{86.87\%}&\textbf{81.12\%}\\
 
\bottomrule
\end{tabular}
\end{table}

\begin{table}

\caption{
 {Comparison of classification accuracy of proposed defenses with other defense strategies, under various attacks on DGCNN and ModelNet40 datasets.}}
\centering
\tiny
\label{table:DGCNN}
\begin{tabular}{llllllll}
\toprule

\bf Defenses  & \multicolumn{7}{|c}{\bf Attacks}\\
\midrule
& Clean & Shift-L2 \cite{xiang2019generating} & Add-CD \cite{xiang2019generating} & Add-HD \cite{xiang2019generating} & Shift-KNN \cite{tsai2020robust} & Drop-100 \cite{zheng2019pointcloud} & Drop-200 \cite{zheng2019pointcloud}\\
 
No-defense & 91.22\% & 0.00\%&1.65\%& 1.57\%&19.23\%&74.54\%&56.53\%\\
 
SRS \cite{yang2021adversarial} & 81.53\% & 51.06\%&63.79\%&43.39\%&41.20\%&50.06\%&23.79\%\\
 
SOR \cite{zhou2019dup} & 88.67\% & 76.61\% &72.49\%&63.81\%&55.93\%&64.59\%&58.99\%\\
 
DUP-Net \cite{zhou2019dup} & 54.05\% & 41.98\%&44.75\%&33.45\%&35.41\%&44.19\%&36.21\%\\
 
If-Defense \cite{wu2020if} & 89.17\% & \textbf{85.49\%}&\textbf{84.15\%}&\textbf{72.88\%}&82.28\%&83.41\%&73.31\%\\
 
Adv Training (Shift-L2) \cite{liu2019extending} & 90.18\% & 13.21\%&6.45\%&6.41\%&15.75\%&75.42\%&54.97\%\\
 
LPF1-Proposed & 91.38\% & 78.97\%&72.16\%&68.52\%&84.04\%&86.99\%&80.19\%\\
 
LPF2-Proposed & \textbf{93.29\%} & 78.69\%&73.63\%&68.68\%&\textbf{85.53\%}&\textbf{88.65\%}&\textbf{82.41\%}\\
 
\bottomrule
\end{tabular}
\end{table}

\begin{table}

\caption{
 {Comparison of classification accuracy of proposed defenses with other defense strategies, under various attacks on PointNet and ScanObjectNN datasets.}}
\tiny
\centering
\label{table:ScanObjectnn in PointNet}
\begin{tabular}{llllllll}
\toprule

\bf Defenses  & \multicolumn{7}{|c}{\bf Attacks}\\
\midrule
& Clean & Shift-L2 \cite{xiang2019generating} & Add-CD \cite{xiang2019generating} & Add-HD \cite{xiang2019generating} & Shift-KNN \cite{tsai2020robust} & Drop-100 \cite{zheng2019pointcloud} & Drop-200 \cite{zheng2019pointcloud}\\
 
No-defense &\textbf{79.17\%} &0.00\%&0.00\%&0.00\%&20.14\%&64.37\%&54.56\%\\
 
SRS \cite{yang2021adversarial}&79.35\% &67.30\%&65.06\%&50\%&71.77\%&65.23\%&56.11\%\\
 
SOR \cite{zhou2019dup}&78.49\% &75.73\%&77.45\%&74.01\%&75.22\%&67.47\%&57.49\%\\
 
DUP-Net \cite{zhou2019dup}&73.67\% &74.70\%&77.28\%&71.43\%&74.18\%&67.99\%&58\%\\
 
If-Defense \cite{wu2020if}&76.41\% &76.00\%&76.24\%&\textbf{75.73\%}&74.52\%&67.46\%&60.92\%\\
 
%
LPF1-Proposed&75.56\% &76.08\%&76.94\%&74.70\%&75.56\%&\textbf{74.01\%}&66.78\%\\
 
LPF2-Proposed&79.00\% &\textbf{77.28\%}&\textbf{77.97\%}&71.94\%&\textbf{76.93\%}&71.94\%&\textbf{68.50\%}\\
 
\bottomrule
\end{tabular}

\end{table}

\begin{table}

\caption{
 {Comparison of classification accuracy of proposed defenses with other defense strategies, under various attacks on PointNet and ShapeNet datasets.}}
\tiny
\centering
\label{table:ShapeNet in PointNet}
\begin{tabular}{llllllll}
\toprule 

\bf Defenses  & \multicolumn{7}{|c}{\bf Attacks}\\
\midrule
& Clean & Shift-L2 \cite{xiang2019generating} & Add-CD \cite{xiang2019generating} & Add-HD \cite{xiang2019generating} & Shift-KNN \cite{tsai2020robust} & Drop-100 \cite{zheng2019pointcloud} & Drop-200 \cite{zheng2019pointcloud}\\

No-defense &94.16\% &0.00\%&0.00\%&0.00\%&18.51\%&86.36\%&80.03\%\\

SRS \cite{yang2021adversarial}&93.67\%&90.75\%&87.5\%&85.55\%&91.88\%&87.34\%&81.17\%\\

SOR \cite{zhou2019dup}&94.32\% &93.02\%&93.83\%&93.18\%&92.69\%&87.34\%&82.14\%\\

DUP-Net \cite{zhou2019dup}&91.07\% &92.86\%&93.18\%&92.86\%&92.53\%&88.31\%&82.63\%\\

If-Defense \cite{wu2020if}&93.34\% &\textbf{93.51\%}&93.34\%&93.18\%&\textbf{92.86\%}&89.12\%&85.71\%\\

%
LPF1-Proposed &91.88\%&90.26\%&90.75\%&91.23\%&90.1\%&90.1\%&86.04\%\\

LPF2-Proposed&\textbf{94.48\%} &93.34\%&\textbf{94.32\%}&\textbf{93.83\%}&92.21\%&\textbf{91.07\%}&\textbf{87.50\%}\\
 
\bottomrule
\end{tabular}

\end{table}

\subsection{Classification Results}
\label{Experiment_Comparison of classification accuracy}




Tables \ref{table:PointNet} to \ref{table:ShapeNet in PointNet} illustrate the point cloud classification accuracy of the proposed defenses and other defense strategies on various attacks. The classification accuracy of defenses is shown as a percentage of the correctly classified test point clouds. A higher classification accuracy in the victim model indicates that the defense is more effective. 





According to the results shown in Tables \ref{table:PointNet} to \ref{table:ShapeNet in PointNet}, the three SRS, SOR, and DUP-NET defenses are effective against point add (add-CD, add-HD) and point shift (Shift-L2) attacks. The mechanism of these defenses allows them to eliminate out-surface perturbation points effectively. In this regard, KNN reduces the accuracy of these defenses as an On-surface perturbation attack (which keeps the perturbation points almost on the surface). Nevertheless, accuracy of defenses still makes sense. The problem occurs when there is a point drop attack and no point to remove. In such attacks, the model accuracy decreases sharply; however, DUP-NET has improved the performance by combining SOR and UPSampler networks by around 4\%. But, because upsampling increases the number of points that are close to the input points, the defense method cannot resist when the number of dropped points becomes too large. Consequently, the add, shift, and drop attacks can be improved with the proposed defense methods.
If-Defense combines SOR and resamples points from a mesh. Also, in order to improve the accuracy of drop attacks and other attacks on the surface, it defines two loss functions to preserve the point geometry and point distribution on the surface. Compared to IF-Defense, the proposed method improves the model accuracy by an average of  2\% on attacks, as shown in Table \ref{table:PointNet}.

Adversarial training (with Shift-L2) trains the models using both original training data and adversarial one (generated by Shift-L2 attacks) \cite{liu2019extending}. Adversarial training (with PAGN) trains the models on a mixture of original training and adversarial data (generated by adaptive attacks) \cite{liang2022pagn}. Adaptive attacks are designed to cover all types of attacks. In fact, these defenses draw the model's attention to the high-frequency components of data (attacks). 
Note that the irregular structure of point clouds can lead to performance instability in such defenses. Table S1 illustrates the summary of methods.

Both $LPF1$ and $LPF2$ use $D_{LP}$ for training. In $D_{LP}$, high-frequency information (which in most adversarial examples is attacked) is removed. For example, a drop attack typically generates adversarial examples by removing corner points and edges. On the other hand, $LPF1$ and $LPF2$, during the training phase, have learned the data whose high-frequency information has been removed. Therefore, these attacks are more likely to be categorized correctly. As shown in Table \ref{table:PointNet}, $LPF2$ has higher accuracy than state-of-the-art defense (\cite{wu2020if}), for about 4\% in both Drop-100 and Drop-200 attacks.

In Table \ref{table:PointNet}, in which the victim model is PointNet, the proposed $LPF2$ method outperforms all defense approaches in all studies attacks except for the Shift-KNN attack for which the IF-Defense approach achieves a higher performance in about $0.9\%$. Table \ref{table:PointNet++} shows the same leading performance of the $LPF2$ method in the PointNet++ model. However, in this case the proposed method performs better in the Shift-KNN attack, but has around $1\%$ lower accuracy in Shift-L2 attack compared to the state-of-the-art approach (i.e., IF-Defense.). It is valuable to note that the $LPF2$ approach has higher accuracy in about 3\%, 1\%, 1\%, 2\%, and 2\% in Add-CD, Add-HD, Shift-KNN, Drop100, and Drop200 attacks than the state-of-the-art approach respectively \cite{wu2020if}.
Similarly, the results for the DGCNN model are also reported in Table \ref{table:DGCNN}. The proposed method outperforms the state-of-the-art approach in Shift-KNN, Drop100, and Drop200 by 3\%, 5\%, and 9\%, respectively. The proposed method has an acceptable performance on Shift-L2, Add-CD, and Add-HD compared to other defense methods except for the IF-Defense approach. Note that the results of the first proposed method ($LPF1$) are also reported in all tables, and they outperform most defense approaches. The best performance is achieved with the proposed $LPF2$ defense, as described above.

We have added the results on two other datasets (ScanObjectNN and ShapeNet) in addition to the famous ModelNet40 dataset to complete the experiments. The proposed method outperforms the existing defense methods in these two additional datasets as well, as shown in Tables \ref{table:ScanObjectnn in PointNet} and \ref{table:ShapeNet in PointNet} (in which PointNet is the victim model).


\begin{figure}
\includegraphics[width=0.5\textwidth]{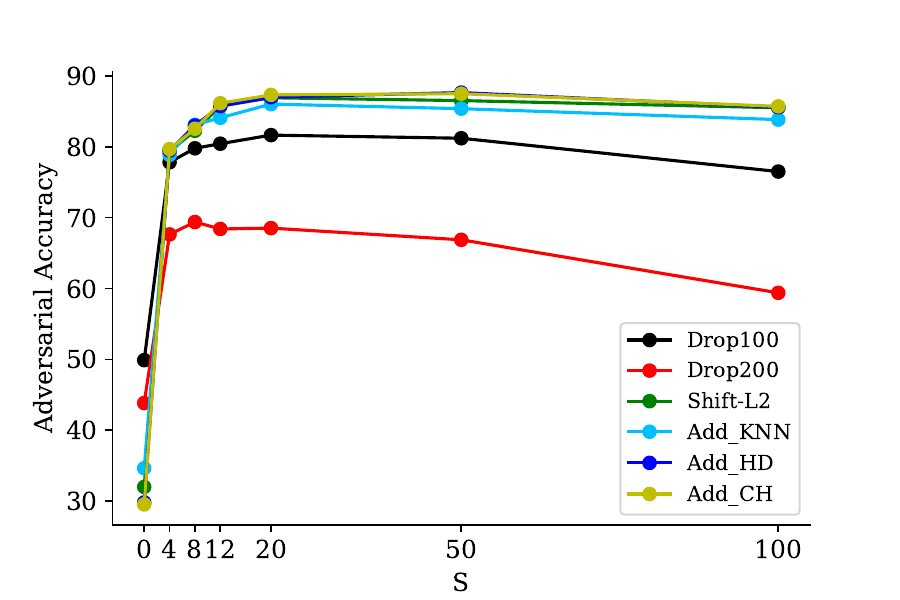}
\centering
\caption{ {Effect of different values of $S$. $LPF2$ results of PointNet model on ModelNet40 dataset. Results are tested on six adversarial attacks. Numbers on horizontal axis indicate the degree of low pass versions with different $S$. }
}
\label{Fig8}
\end{figure}



\begin{figure}
\centering
\includegraphics[width=0.7\textwidth]{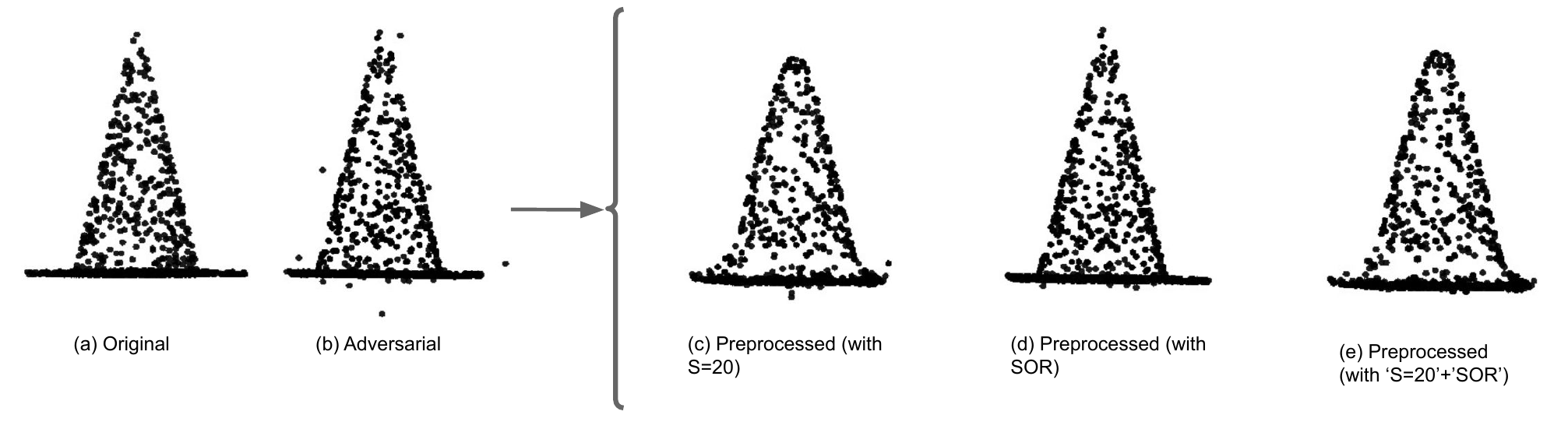}

\caption{ {Visualization of different preprocessors on an adversarial point cloud sample.}
(a) Illustrate original point cloud and others illustrate adversarial point clouds when (b) generated with Add-CD attack \cite{xiang2019generating}, (c) Preprocessed with $S$ = 20, (d) preprocessed with SOR, and (e) Preprocessed with both SOR and $S$ = 20. As expected, (e) has the lowest outliers.}
\label{Fig6}
\end{figure}

As a preprocessing step for the add and shift point attacks, SOR is first applied to remove the outliers. Then, the low-frequency information of data is retained. The results show that the combination of two utterly different denoising mechanisms (SOR and $D_{LP}$) helps to boost the model robustness. In Fig \ref{Fig6}, the Add-CD attack has been applied to the original point cloud. Then, three different methods for removing outliers are tested. Firstly, $D_{LP}$ with S = 20 removes most of the outliers, except for one point at the bottom and one point in the rightmost side of the object. Secondly, SOR method that is the best in dropping outliers, removes the outliers. As seen in this figure, except for the few points at the bottom of the object that are very close to the object, SOR removes the outliers, effectively. Finally, the combination of $D_{LP}$ and SOR takes the advantage of both methods and therefore the resulting object contains the least number of outliers.

\begin{figure}
\includegraphics[width=\textwidth]{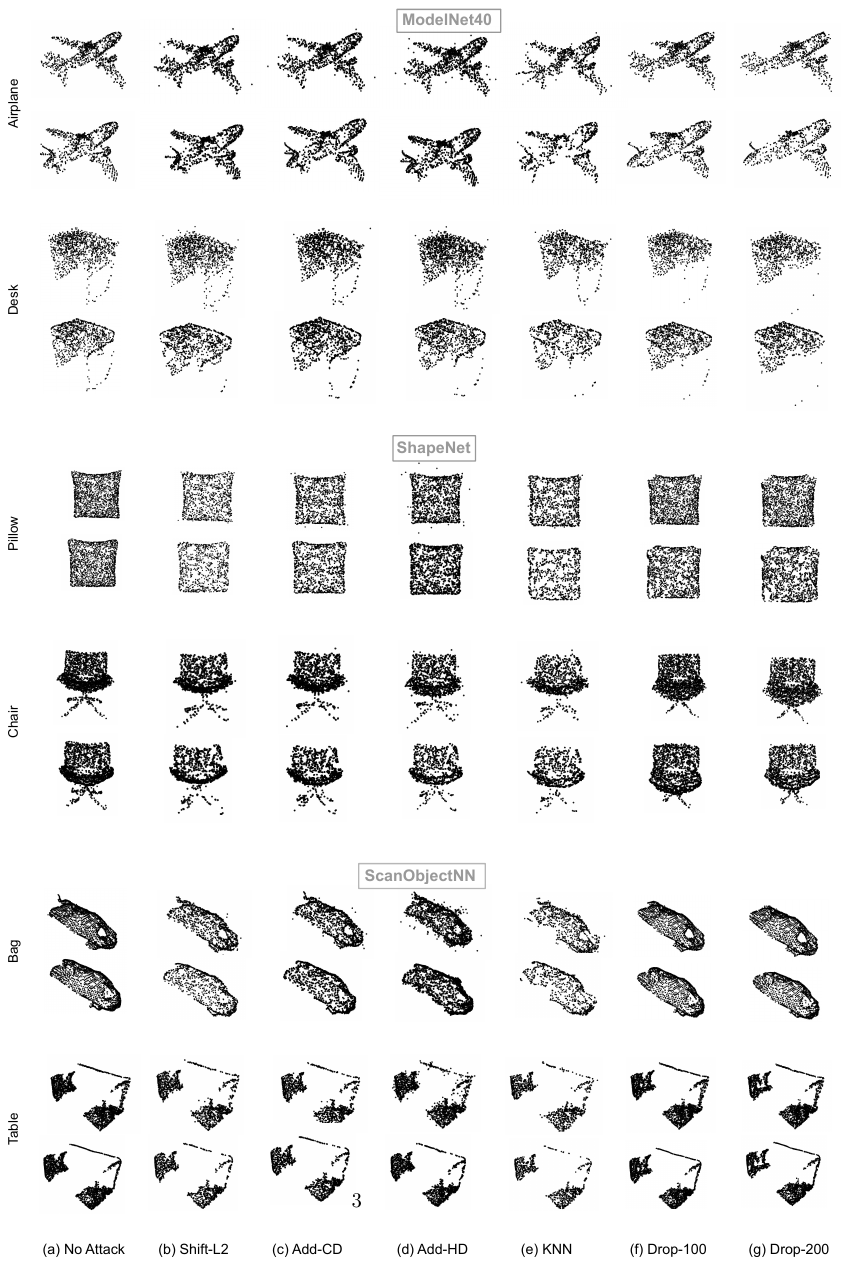}
\centering
\caption{ Effect of proposed low pass preprocessing on point cloud samples of three datasets Modelnet40, ShapeNet, and ScanobjectNN for different attacks. 
 [Odd rows: Point clouds before preprocessing. Even rows: Point clouds after low pass filtering with $S = 20$. Preprocessing in (b), (c), and (d) is performed with a combination of low pass filtering and SOR.]}
\label{Fig7}
\end{figure}

Even rows in Fig \ref{Fig7} illustrate the same preprocess as panel (e) in Fig \ref{Fig6}, on various adversarial attacks for point cloud examples. Odd rows show the non preprocessed point clouds.


\subsection{Ablation Study}

Fig. \ref{Fig8} shows the impact of different low pass frequencies on the performance of $LPF2$, by setting standard deviations $S$ to 0, 4, 8, 12, 20, 50, and 100. Setting $S$ to different values is equivalent to generating low-frequency versions of $D_{LP}$ and training and testing the model with them. In this experiment, six adversarial attacks are used to investigate the model accuracy. As shown in Fig. \ref{Fig8}, the accuracy starts to increase from $S$ = 0 and peaks at $S$ = 20 and then decreases. However, there are exceptions. For example, in the drop-200 attack, the accuracy peaks at $S$ = 8 and increases again at $S$ = 20.
The amount of $S$ that the model is trained on can significantly affect the model's performance on adversarial attacks. If the amount of $S$ is too small (0 or 4), the object's appearance gets closer to the sphere (as seen in Fig. \ref{Fig1}). Also, most of the high frequencies' information is not presented. In such cases, objects from different classes are not different, even in appearance. For $S$=20, both the object appearance and the amount of high frequencies' information are acceptable on average. The more $S$ increases, the object's appearance gets closer to the original point cloud. However, higher frequencies' information in the objects undermines the model's robustness to the adversarial examples.

\begin{figure}
\centering
\includegraphics[width=0.5\textwidth]{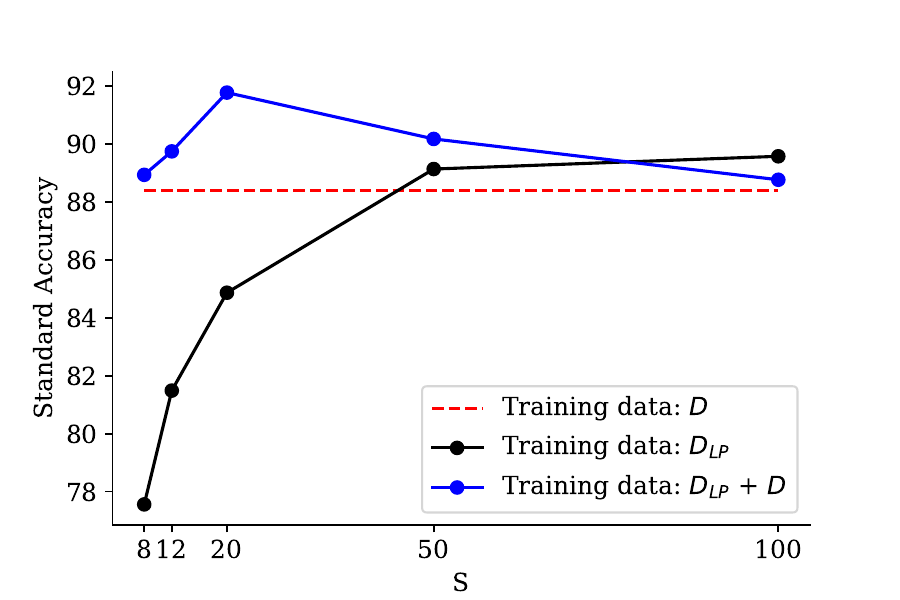}
\caption{Comparison of standard accuracy with three different training data on PointNet model. Three approaches are tested on ModelNet40 test data. Numbers on horizontal axis indicate the degree of low pass versions with different $S$.}

\label{fig:Robustness}
\end{figure}

\subsection{Robustness}
Although the focus of this paper is on the adversarial robustness of models, model training with $D_{LP}$ can also improve the robustness performance. It should be noted that adversarial robustness refers to improving model robustness on adversarial examples. But, the term robustness refers to improving the model's performance on the original inputs. Fig. \ref{fig:Robustness} shows the standard accuracy in three different ways. In Method 1, the model is trained with $D$ (the original ModelNet40 dataset). In Method 2, the model is trained with $D_{LP}$ (the low pass version of the ModelNet40 dataset). In Method 3, the model is trained by a combination of $D_{LP}$ and $D$. Also, in Methods 2 and 3, the parameter $S$ in $D_{LP}$ is set to 0, 4, 8, 12, 20, 50, and 100. The accuracy of these three methods on original test data is shown in Fig. \ref{fig:Robustness}. Note that all the three methods are evaluated on the original ModelNet40 test dataset. The only difference is in the training data. It is seen in Fig. \ref{fig:Robustness} that the accuracy of Method 3 with all $S$ values and the accuracy of Method 2 with the $S$ values of 50 and 100 are higher than the standard accuracy. This means that model training with Methods 2 and 3 can enhance robustness compared to model training with Method 1. 

In Method 2, the accuracy increases with increasing the parameter $S$. At first (in the lower $S$ values), the appearance of the objects (as initially shown in Fig. \ref{Fig1}) is far from the original data and the model training is done only with these objects.
In these cases  ($S = 8,\, 12,\, 20$), it would be reasonable for the model to predict lower accuracy than the standard (Method 1). As $S$ increases ($S = 50, \, 100$), it is observed that the accuracy grows higher than Method 1. The appearance (refer to Fig. \ref{Fig1}) is more similar to the original one. Also, unlike Method 1, the training data still does not contain all high-frequency information, resulting in a better performance. In Method 3, adding $D_{LP}$ to $D$ increases the accuracy in all different $S$ values. It seems that injecting low-frequency data ($D_{LP}$) alongside $D$ can robust the model even against the original data. More interestingly, this accuracy peaks at $S = 20$. However, by increasing $S$, the accuracy remains higher than Method 1. In fact, this analysis shows that by removing tiny perturbations from the training data in specific directions (high frequency) can lead to boost the model robustness.

\section{Conclusion}
\label{sec:Conclusion}
In this paper, a novel perspective was used to analyze adversarial perturbations. As most of the perturbations occur in high-frequency components, to select the appropriate range of high-frequency components, two experiments were conducted. According to the first experiment, Gaussian filters are more suitable than box filters for achieving cut-off frequencies. In the second experiment, by varying the Gaussian filter standard deviation on different values of  $S$, it was concluded that $S= 20$ is an admissible approximation for the cut-off frequency. These results led to the development of new defense methods. Consequently, the LPF-Defense is proposed to improve the performance of models by suppressing high-frequency information in the training phase. Results showed that the LPF-Defenses increased model accuracy compared to state-of-the-art defense methods in six different adversarial attacks. It also demonstrated a positive impact on the models’ robustness against original point clouds, in addition to enhancing robustness against adversarial point clouds.
It should be noted that due to the multilayer nonlinear structure of DNNs, obtaining an accurate description of them is close to impracticable. This complex structure makes model decisions dependent on numerous features. In this paper, features are studied only from a frequency perspective, which may overlook other aspects of the model that are also important.

\section{Acknowledgments}
The authors would like to thank 
Professor Ivan V. Bajić and Dr. Chinthaka Dinesh for the helpful discussions.

\bibliographystyle{unsrt}  
\bibliography{References}

\clearpage
\appendix
\label{sec:6}

\section{}

\paragraph*{Table A}
\label{S33_Table}
 Table S1 compares the various defense methods. The column entitled ``Type'' specifies the general technique that is used in each defense method. In the column entitled "Method", a brief description of the defense method is introduced to give a better perspective. Finally, the column "strength" indicates each method's level of success on three main attack mechanisms. Note that the information in this column is general, and the reader might refer to tables \ref{table:PointNet} - \ref{table:ShapeNet in PointNet} for a detailed comparison. Note that the defense methods of type \textit{modified input} mainly focus on outlier removing and generating a uniform distribution of the points in the point cloud. On the other hand, the defense methods of type \textit{modified training} train the model with the original data as well as the attack data (which contains outliers). Finally, the proposed methods take advantage of both types by training with the original data while removing the high frequency information. This covers the disadvantage of the \textit{modified training} idea due to removal of outliers considered as the high frequency components.

\begin{landscape}
\begin{table}[!ht]
{\bf Comparison of various defense methods with proposed methods in terms of type, methodology, and strength of their defense.}
\centering\small\small\small
\label{S2_Table}

\begin{tabular}{p{2.2cm}|p{1.1cm}|p{13cm}|cccccc}
\hline 
\bf Defenses  & \multicolumn{1}{c|}{\bf Type} & \multicolumn{1}{c|}{\bf Method} & \multicolumn{3}{c}{\bf strength} \\

& \multicolumn{1}{c|}{} & \multicolumn{1}{c|}{} & \multicolumn{1}{c|}{Point add} & \multicolumn{1}{c|}{Point shift} & \multicolumn{1}{c}{Point drop} \\

\hline 
\multirow{1}{*}{SRS \cite{yang2021adversarial}} & input &  Outlier removal is done by randomly dropping points. &*** & ** & *\\

\hline
\multirow{1}{*}{SOR \cite{zhou2019dup}} & input & Outlier removal is done based on the point distance to its k-nearest neighbors.&***&**&*\\

\hline
\multirow{2}{*}{DUP-Net \cite{zhou2019dup}} & input &  Outlier removal is done by SOR &*** & *** & ** \\&   & Uniform distribution of points is achieved using the upsampler network.&&&\\

\hline
\multirow{2}{*}{If-Defense \cite{wu2020if}} & input &  Outlier removal is done based on limiting the point perturbation with new loss definition + SOR &**** & **** & *** \\&   & Uniform distribution of points is achieved by imposing penalty on close points using new loss definition. &&&\\

\hline
\multirow{2}{*}{} Adv Training& Training & Trains the model with original data and Shift-L2 attacked data. &** & * & ***  \\(Shift-L2) \cite{liu2019extending}&   & training is done with add adversarial outliers&&&\\

\hline
\multirow{2}{*}{}Adv Training & Training & Trains the model with original data and adaptive attacked data. &** & *** & ** \\(PAGN) \cite{liang2022pagn}&   & training is done with add adversarial outliers with outlier-inject-module&&&\\

\hline
\multirow{3}{*}{LPF1-Proposed} & Training & Trains the model with information of Low-Freq data. &**** & **** & **** \\& /input  & Training is done by removing high-freq information as outliers. &&&\\&   & Outlier removing is done by LF+SOR. &&&\\

\hline
\multirow{3}{*}{LPF2-Proposed} & Training & Trains the model with original data and the information of Low-Freq data. &**** & **** & *****  \\&  /input & Training is done by removing high-freq information as outliers.&&&\\&   & Outlier removing is done by LF+SOR. &&&\\
 
\hline
\end{tabular}
\end{table}
\end{landscape}

\clearpage
\section{}


\paragraph*{Table B}
\label{S22_Table}
 In order to analyze the performance, a confusion matrix is calculated, and then the accuracy, f1-score, precision, and recall criteria are derived from it. Table S2 displays that the proposed method performs well with all three different metrics in the ShapeNet dataset on the PointNet model.

\begin{table}[!ht]

{\bf Comparison of classification’s f1-score, precision, and recall measures of proposed defenses with other defense strategies, under various attacks on PointNet and ShapeNet datasets.[F1-scores, precision, and recall results for each defense are shown in the first, second, and third row, as a percentage, respectively.]}
\tiny 
\centering
\label{S222_Table}
\begin{tabular}{llllllll}

\hline 
\bf Defenses  & \multicolumn{7}{c|}{\bf Attacks}\\
\hline
\hline
& Clean & Shift-L2 \cite{xiang2019generating} & Add-CD \cite{xiang2019generating} & Add-HD \cite{xiang2019generating} & Shift-KNN \cite{tsai2020robust} & Drop-100 \cite{zheng2019pointcloud} & Drop-200 \cite{zheng2019pointcloud}\\
\hline 

\multirow{3}{*}{SRS \cite{yang2021adversarial}} &\textbf{87.7\%}&74.5\%&68.6\%&65.2\%&84.9\%&72.2\%&57.8\%\\
&84\%&76.2\%&72.6\%&70.6\%&81.9\%&73.3\%&62.9\%\\
&\textbf{96.2\%}&75\%&67.4\%&63.5\%&89.4\%&72.8\%&56.5\%\\
\hline 

\multirow{3}{*}{SOR \cite{zhou2019dup}}
&88.8\%&83.3\%&85.3\%&84\%&\textbf{86\%}&69.5\%&60.2\%\\
&\textbf{85}\%&78.9\%&80.8\%&79.9\%&\textbf{83\%}&71\%&64.8\%\\
&96.3\%&92.16\%&\textbf{95.7\%}&91.8\%&90.4\%&69.6\%&58.6\%\\
\hline 

\multirow{3}{*}{DUP-Net \cite{zhou2019dup}}
&77.6\%&81.5\%&83.7\%&85.3\%&84.2\%&69.7\%&60.8\%\\
&76.1\%&78.16\%&81.1\%&81.8\%&82\%&70.7\%&64.8\%\\
&85.4\%&91.6\%&91.7\%&\textbf{93.7}\%&89.5\%&70.7\%&59.6\%\\
\hline 

\multirow{3}{*}{If-Defense \cite{wu2020if}}
&85.6\%&85.4\%&84.2\%&84.3\%&82.05\%&75.5\%&64.8\%\\
&82.2\%&81.7\%&82\%&81.5\%&77.8\%&75.9\%&68.2\%\\
&92.7\%&\textbf{93.8}\%&92.2\%&93.4\%&\textbf{93.3}\%&76.6\%&63.1\%\\
\hline 

\multirow{3}{*}{LPF1-Proposed}
&74.2\%&75.3\%&77.4\%&77.6\%&74.9\%&76.9\%&66\%\\
&72.9\%&72.8\%&74.3\%&74.8\%&71.8\%&77\%&70.9\%\\
&77.9\%&84.7\%&90\%&87.1\%&82.1\%&85.8\%&69\%\\
\hline 

\multirow{3}{*}{LPF2-Proposed}
&86.6\%&\textbf{88.6}\%&\textbf{88.5}\%&\textbf{89.7}\%&83.6\%&\textbf{84.8\%}&\textbf{78.1\%}\\
&82.9\%&\textbf{86.3\%}&\textbf{85.7\%}&\textbf{87.6\%}&82.9\%&\textbf{82.7\%}&\textbf{79.4}\%\\
&93.4\%&91.5\%&93.7\%&92.5\%&88.01\%&\textbf{88.7\%}&\textbf{77.5\%}\\
\hline

\hline
\end{tabular}

\end{table}

\clearpage
\section{}

\paragraph*{ Table C}
\label{S3_Table}
 Table S3 compares the computational cost for the preprocessing step in different defense methods. It shows the average inference time of one adversarial point cloud (shift-l2) for several defense methods. While the low pass filtering method proposed in this paper is computationally heavier than SOR, SRS, and DUP-Net defense methods, it beats the state-of-the-art IF-Defense. The experiment is conducted on a Google Colab environment with a NVIDIA Tesla T4 GPU.

\begin{table}[hbt!]

{\bf Computational cost of preprocessing. Time column lists the average inference time required to compute one point cloud.}

\centering
\scriptsize
\label{table:Computational cost}
\begin{tabular}{|c|c|}
\hline 
\bf Defenses  & \multicolumn{1}{c|}{\bf Time (millisecond)} \\
\hline
\hline 
SRS \cite{yang2021adversarial}&0.1ms\\
\hline 
SOR \cite{zhou2019dup}&2.6ms\\
\hline 
DUP-Net \cite{zhou2019dup}&13.66\\
\hline 
If-Defense \cite{wu2020if}&164ms\\
\hline 
LPF-Proposed&73.86ms\\
 
\hline
\end{tabular}

\end{table}

\end{document}